\documentclass[conference]{IEEEtran}
\IEEEoverridecommandlockouts

\usepackage{cite}
\usepackage{amsmath,amssymb,amsfonts}
\usepackage{algorithmic}
\usepackage{graphicx}

\usepackage{textcomp}
\usepackage{xcolor}
\usepackage{bibentry}
\usepackage{tabularx} 
\usepackage{booktabs} 
\usepackage{multirow} 
\usepackage{hyperref}

\usepackage{subfigure}

\def\BibTeX{{\rm B\kern-.05em{\sc i\kern-.025em b}\kern-.08em
    T\kern-.1667em\lower.7ex\hbox{E}\kern-.125emX}}

\makeatletter
\newcommand{\linebreakand}{%
  \end{@IEEEauthorhalign}
  \hfill\mbox{}\par
  \mbox{}\hfill\begin{@IEEEauthorhalign}
}
\makeatother

\begin{document}

\title{Multi-Normal Prototypes Learning for Weakly Supervised Anomaly Detection
}

\author{ \IEEEauthorblockN{Zhijin Dong}
\IEEEauthorblockA{\textit{School of Software and Microelectronics} \\
\textit{Peking University}\\
Beijing, China \\
zhijindong@pku.edu.cn}
\and
\IEEEauthorblockN{Hongzhi Liu$^*$\thanks{*Corresponding author.}}
\IEEEauthorblockA{\textit{School of Software and Microelectronics} \\
\textit{Peking University}\\
Beijing, China \\
liuhz@pku.edu.cn}
\linebreakand
\IEEEauthorblockN{Boyuan Ren}
\IEEEauthorblockA{\textit{School of Software and Microelectronics} \\
\textit{Peking University}\\
Beijing, China \\
rby\_tj@stu.pku.edu.cn}
\and
\IEEEauthorblockN{Weimin Xiong}
\IEEEauthorblockA{\textit{School of Computer Science} \\
\textit{Peking University}\\
Beijing, China \\
wmxiong@pku.edu.cn}
\and
\IEEEauthorblockN{Zhonghai Wu\thanks{Under review.}}
\IEEEauthorblockA{\textit{National Engineering Center of Software Engineering} \\
\textit{Peking University}\\
Beijing, China \\
wuzh@pku.edu.cn}

}

\maketitle

\begin{abstract}
Anomaly detection is a crucial task in various domains. Most of the existing methods assume the normal sample data clusters around a single central prototype while the real data may consist of multiple categories or subgroups. In addition, existing methods always assume all unlabeled samples are normal while some of them are inevitably being anomalies. To address these issues, we propose a novel anomaly detection framework that can efficiently work with limited labeled anomalies. Specifically, we assume the normal sample data may consist of multiple subgroups, and propose to learn multi-normal prototypes to represent them with deep embedding clustering and contrastive learning. Additionally, we propose a method to estimate the likelihood of each unlabeled sample being normal during model training, which can help to learn more efficient data encoder and normal prototypes for anomaly detection. Extensive experiments on various datasets demonstrate the superior performance of our method compared to state-of-the-art methods. Our codes are available at: \href{https://github.com/Dongzhijin/MNPWAD}{\texttt{https://github.com/Dongzhijin/MNPWAD}}
\end{abstract}


\section{Introduction}
Anomaly detection aims to identify samples that deviate significantly from the general data distribution, known as anomalies or outliers \cite{weakly-survey23}. These anomalies often indicate potential problems or noteworthy events that require attention \cite{review21}. The significance of anomaly detection spans a variety of critical domains, such as network security \cite{NetworkAnomaly20}, Internet of Thing (IoT) security \cite{IoTsecurity22}, and financial fraud detection \cite{FraudAnomaly18}. In each of these areas, effective anomaly detection methods are essential due to the substantial impact of anomalies, which can lead to severe consequences such as security breaches\cite{cybersecurity24}, financial losses\cite{FinancialIoT24}, compromised privacy\cite{FederatedAD24}, and even risks to human health and safety\cite{healthcare24,RoadNetwork23}.

In real-world scenarios, obtaining labeled data for anomaly detection is often challenging and costly \cite{weakly-survey23,ADGym23,time-series-anomaly-SelfSupervised24}. Although unsupervised methods\cite{time-series-anomaly-Graph24,time-series-anomaly-DAEMON21} can bypass the need for extensive labeling, they struggle to achieve optimal performance without knowing what true anomalies look like. This limitation frequently results in the misidentification of many normal data points as anomalies \cite{PreNet23}. To address these challenges, \textbf{weakly supervised anomaly detection} methods have emerged \cite{DevNet19,RoSAS23}. These methods leverage a small portion of labeled anomalous data in conjunction with a large amount of unlabeled data, primarily composed of normal instances. This kind of approach is particularly significant in practical applications where we can only label a small fraction of anomalies. 
\begin{figure}[t]
  \centering
  \includegraphics[width=0.95\columnwidth]{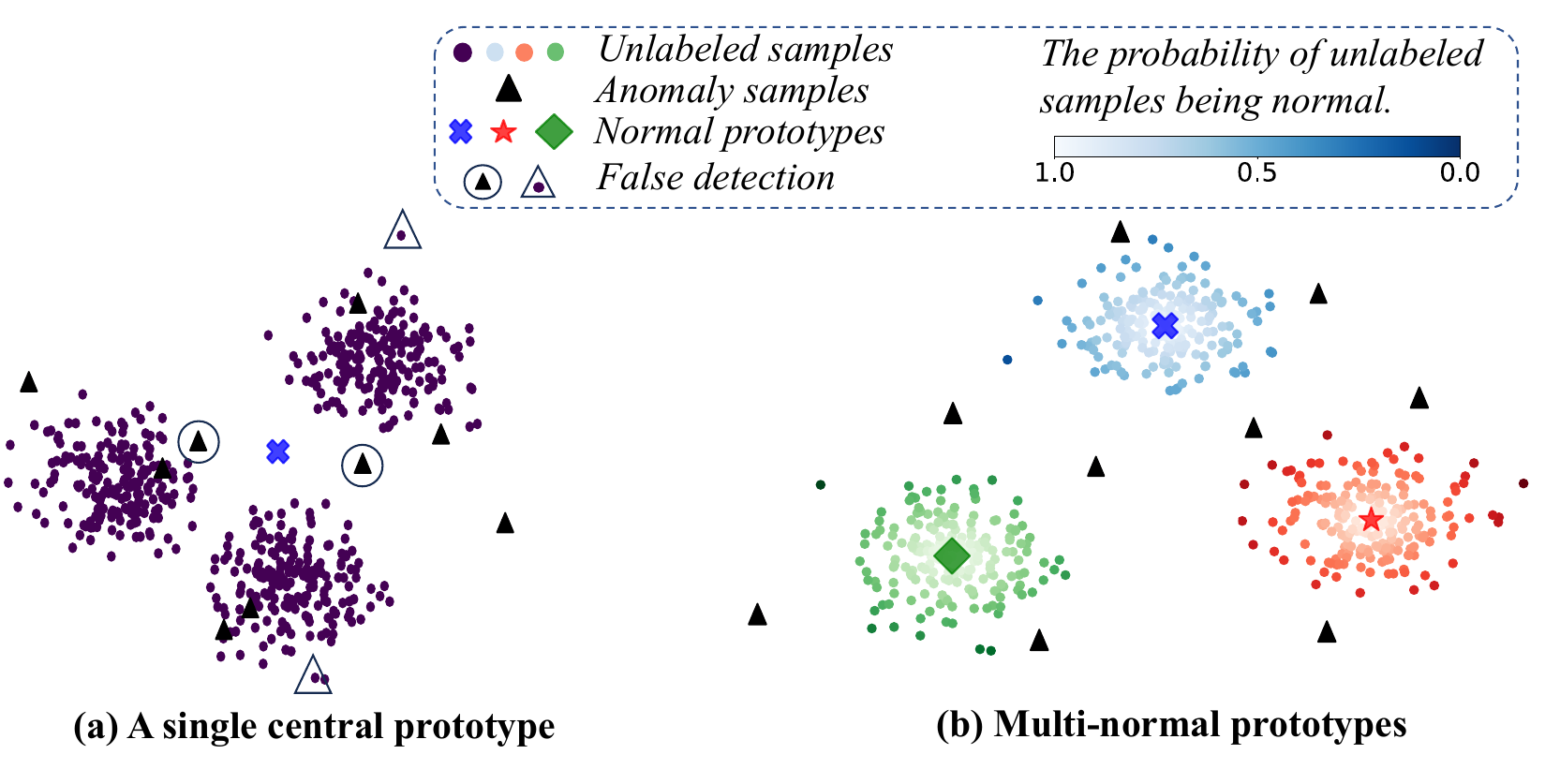}
  \caption{Motivations of the proposed multi-normal prototype learning (b). Compared with single-normal prototype learning (a), multi-normal prototype (b) can more effectively capture the diversity of normal data and estimate the probability of unlabeled samples being normal to resist anomaly contamination.}
  \label{fig:Multi normal prototypes}
\end{figure}

However, existing anomaly detection methods still face several significant challenges. First, most of the existing anomaly detection methods \cite{DeepSVDD18,DeepSAD20,DevNet19} assume that normal data clusters around a single central prototype. This assumption oversimplifies real-world scenarios where normal data often comprises multiple categories or subgroups, as illustrated in Fig. \ref{fig:Multi normal prototypes}. Such methods struggle to capture the complexity and diversity of normal data, leading to potential misclassification of normal samples as anomalies \cite{DPAD24}.
Second, existing methods always assume all unlabeled data are normal while they inevitably contain some anomalous samples. Due to the assumption, many studies \cite{FeaWAD21,DPAD24,DevNet19} directly employ large amounts of unlabeled data as inliers for model training to identify the distribution of normal data patterns. These methods are vulnerable to the presence of occasional anomalies (i.e., anomaly contamination)\cite{DeepAnomaly23} within the unlabeled data, and the detection accuracy rapidly declines as the proportion of mixed-in anomalies increases \cite{PreNet23,RoSAS23}. 

To address the problems stated above, we propose a reconstruction-based multi-normal prototype learning framework for weakly supervised anomaly detection. Different from existing reconstruction-based anomaly detection methods \cite{FeaWAD21}, we treat the normal samples and anomalous samples differently during the latent representation learning with consideration of the likelihood of each unlabeled sample being normal. To better estimate the likelihood of normal and detect anomalies, we propose to learn multiple normal prototypes in the latent space with deep embedding clustering and contrastive learning. Finally, we compute a comprehensive anomaly score with consideration of both the information of sample reconstruction and multiple normal prototypes.

The main contributions of this paper can be summarized as follows:
\begin{itemize}
\item  We propose a novel anomaly detection framework that combines reconstruction learning with multi-normal prototype learning. Extensive experiments across various datasets demonstrate that our method significantly outperforms state-of-the-art methods.
\item  We propose to build multiple normal prototypes with deep embedding clustering and contrastive learning to better model the distribution of normal data for anomaly detection. 
\item  We propose to estimate and take into consideration the likelihood of each unlabeled sample being normal during model training, which can enhance resistance to anomaly contamination and more effectively detect unseen anomaly classes.
\end{itemize}

\section{Related Work}
\subsection{Anomaly Detection Based on Prototype Learning}
One-class classification methods \cite{DOC18} are a fundamental category of traditional anomaly detection techniques. These methods are trained solely on normal data and identify anomalies by detecting deviations from a central prototype. Examples include One-Class SVM (OC-SVM) \cite{One-classSVM}, Support Vector Data Description (SVDD) \cite{SVDD}, Deep SVDD \cite{DeepSVDD18}, and DeepSAD \cite{DeepSAD20}, all of which assume that normal samples cluster around a single prototype in feature space, with anomalies located at the periphery. However, some studies have shown that normal data frequently consists of multiple categories or subgroups \cite{DPAD24}, rendering a single prototype insufficient \cite{PRNAD2023}. To address these limitations, multi-prototype learning approaches have been developed to better represent the diversity within normal data, thereby enhancing anomaly detection capabilities. This concept has been applied across various domains, such as video anomaly detection \cite{park2020}, image anomaly detection \cite{ProtoAD2024}, and time series analysis \cite{PETAD2023,time-series-anomaly23,time-series-anomaly22}. These advancements highlight the potential of multi-prototype learning, particularly in the context of tabular data for weakly supervised anomaly detection. 
By capturing the diversity inherent in normal data through multiple prototypes, our proposed method aims to more effectively detect anomalies that would otherwise be missed by single-prototype models.

\begin{figure*}[th]
  \centering
  \includegraphics[width=0.8\textwidth]{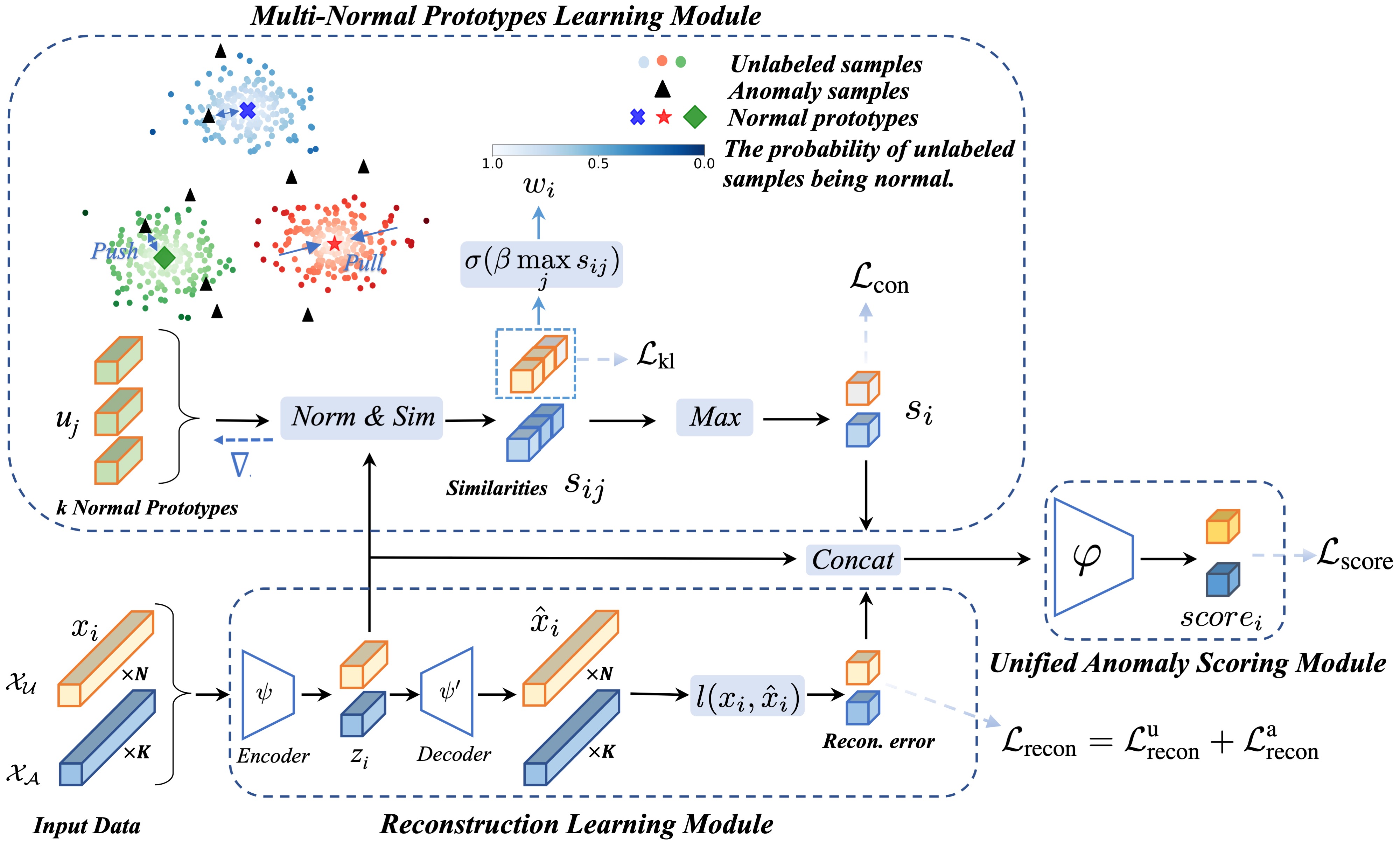}
\caption{Overview of the proposed model. The input \(x_i\) from a labeled anomaly set \(\mathcal{X_A}\) and an unlabeled set \(\mathcal{X_U}\) are represented by blue and yellow shades, respectively. The output \(score_i\) indicates the likelihood of a sample being an anomaly.}
  \label{fig:model}
\end{figure*}

\subsection{Weakly Supervised Anomaly Detection}
Weakly supervised anomaly detection has emerged as a solution to the limitations of purely unsupervised methods, which struggle to identify what anomalies look like due to the lack of labeled examples \cite{PreNet23}. In weakly supervised settings, a small number of labeled anomalies are combined with a large amount of unlabeled data to enhance detection performance. Existing weakly supervised methods face notable challenges. Many rely heavily on large amounts of unlabeled data to identify normal data patterns, making them susceptible to contamination by occasional anomalies within the unlabeled dataset \cite{FeaWAD21,DeepSAD20}. This lack of effective mechanisms to mitigate the impact of these anomalies can significantly degrade detection performance, with accuracy declining rapidly as the proportion of mixed-in anomalies increases \cite{FeaWAD21, DPAD24}. Moreover, a common issue across these methods \cite{PreNet23,RoSAS23,DeepSAD20,DevNet19} is overfitting to the limited labeled anomalies. This over-fitting weakens the generalization capability of the models, especially when the number of known anomalies is small or when the models are confronted with previously unseen anomaly types during the testing phase. As a result, detection performance suffers, highlighting the need for methods that can better leverage the small amount of labeled data while effectively utilizing the distribution patterns within a large pool of unlabeled data.

\section{Methodology}
\subsection{Task Definition}
Let $\mathcal{X} = \{x_i\}_{i=1}^{N+K}$ denote the sample set, which contains two disjoint subsets, $\mathcal{X_U}$ and $\mathcal{X_A}$. Here, $\mathcal{X_U} = \{x_i\}_{i=1}^{N}$ represents a large set of unlabeled samples, and $\mathcal{X_A} = \{x_i\}_{i=N+1}^{N+K}$ (where $K \ll N$) denotes a small set of labeled anomaly samples. Our objective is to learn a scoring function $\phi: \mathcal{X} \rightarrow \mathbb{R}$ that assigns anomaly scores to each data instance, with the larger value of $\phi(x_i)$ indicating a higher likelihood of $x_i$ being an anomaly. 

\subsection{Overview of Our Method}
To detect anomalies in a weakly supervised setting, we propose a novel approach combining reconstruction learning and multi-normal prototype learning, as illustrated in Fig. \ref{fig:model}. It consists of three main components: 1) \textbf{Reconstruction Learning Module} aims to guide the latent representation learning with differentiated treatment to reconstruction errors of normal and anomalous samples; 2) \textbf{Multi-Normal Prototypes Learning Module} aims to model the distribution of normal data better and utilizes deep embedding clustering and contrastive learning to build multiple normal prototypes for anomaly detection; 3) \textbf{Unified Anomaly Scoring Module} aims to compute a comprehensive anomaly score with consideration of both the information of sample reconstruction and multiple normal prototypes. The detailed design of each component is described below.

\subsection{Reconstruction Learning Module}
The primary purpose of this module is to transform the data into a latent space that captures the essential features of each sample, enabling effective differentiation between normal and anomalous samples. These latent representations are subsequently utilized in the Multi-Normal Prototypes Learning Module and serve as one of the inputs to the Unified Anomaly Scoring Module. Unlike traditional reconstruction-based methods\cite{autoencoders17,Autoencoder22}, which aim to minimize the reconstruction error for all samples without considering their labels, our approach incorporates label information and the likelihood of unlabeled samples being normal.

First, we employ an encoder-decoder structure for reconstruction learning. The encoder function \(\psi\) maps an input \(x_i\in \mathbb{R}^D\) to a latent representation \(z_i\in \mathbb{R}^H\):
\begin{equation}
z_i = \psi(x_i; \Theta_{\psi})
\end{equation}
The decoder function \(\psi^\prime\) maps the latent representation \(z_i\) back to a reconstruction \(\hat{x}_i\in \mathbb{R}^D\) of the original input:
\begin{equation}
\hat{x}_i = \psi^\prime(z_i; \Theta_{\psi^\prime})
\end{equation}
where \(\Theta_{\psi}\) and \(\Theta_{\psi^\prime}\) are the parameters of the encoder and decoder neural networks, respectively.

Second, we specifically design a reconstruction loss for weakly supervised anomaly detection. On the one hand, we minimize the reconstruction error to ensure that the latent representations capture the most relevant characteristics of the normal data. For unlabeled samples, the reconstruction loss is defined as:
\begin{equation}
\mathcal{L_{\text{recon}}^{\text{u}}} = \mathbb{E}_{x_i \sim \mathcal{X_U}} [w_i \mathit{l}(x_i, \hat{x}_i)]
\label{eq:recon}
\end{equation}
where \(\mathit{l}\) is a regression loss function like Mean Squared Error (MSE), and \(w_i\) indicates the probability of an unlabeled sample being normal, reducing the impact of anomaly contamination in the training process by assigning them lower weights. The next module will detail how \(w_i\) is derived.

On the other hand, to distinguish normal samples from anomalies, we apply a hinge loss \cite{hingeloss} to ensure that normal samples have a lower reconstruction loss compared to anomalies, maintaining a sufficient margin \(m_1\). Therefore, the reconstruction loss for labeled anomalies is defined as:
\begin{equation}
\mathcal{L_{\text{recon}}^{\text{a}}} =\mathbb{E}_{x_i \sim \mathcal{X_A}} [\max\left(0, m_1-(\mathit{l}(x_i, \hat{x}_i)-\mathcal{L_{\text{recon}}^{\text{u}}})\right)]
\end{equation}
The overall reconstruction loss, combining the losses for both unlabeled data and labeled anomalies, is given by:
\begin{equation}
\mathcal{L_{\text{recon}}} = \mathcal{L_{\text{recon}}^{\text{u}}}+\mathcal{L_{\text{recon}}^{\text{a}}} 
\end{equation}

\subsection{Multi-Normal Prototypes Learning Module}
We assume the normal sample data may consist of multiple categories or subgroups and satisfy multi-modal distribution. Therefore, we propose to learn multiple prototypes to represent the normal data. Specifically, we utilize deep embedding clustering and contrastive learning to learn multiple normal prototypes.

\paragraph{Initialization and Similarity Calculation} 
At the beginning of training, we initialize \( k \) normal prototypes by clustering normal data's latent representations from the pre-train Encoder, denoted as \( \{u_j\in \mathbb{R}^H\}_{j=1}^k \). During the model training and inference phases, we need to calculate the similarity between a given sample $x_i$ and each prototype $u_j$ to determine which category or subgroup the sample belongs to. Specifically, we assume the samples of each category or subgroup satisfy the Student’s $t$-distribution, and calculate the similarity between sample $x_i$ and prototype $u_j$ as follows: 
\begin{equation}
    s_{ij}=\left(1 + \frac{\|z_i - u_j\|^2}{\alpha}\right)^{-\frac{\alpha+1}{2}}
\end{equation}
where $z_i$ denotes the latent representation of sample $x_i$, \(\alpha\) is the degree of freedom of the Student’s t-distribution, $u_j$ denotes the latent representation of the $j$-th normal prototype. To ensure consistency and stability during training, both the normal prototype $u_j$ and the latent representation \( z_i \) are normalized before calculating the similarity.

\paragraph{Weights of Unlabeled Samples} 
In our anomaly detection framework, the weights of unlabeled samples are crucial for adjusting the influence of each sample during the training process. 
The weights, which represent the probability of unlabeled samples being normal, help to mitigate the contamination effect of anomalies within unlabeled data.
We define the weight \(w_i\) for an unlabeled sample \(x_i\) based on its maximum similarity to any normal prototype:
\begin{equation}
    w_i = \sigma(\beta \max\limits_{j} s_{ij})
\end{equation}
where \(\sigma\) is the sigmoid activation function, and \(\beta\) is a scaling parameter that adjusts the sensitivity of the similarity scores.

Unlabeled samples with higher similarity to normal prototypes are assigned greater weights, increasing their influence in the loss function associated with unlabeled data and focusing the model on preserving the characteristics of normal samples. Conversely, unlabeled samples with lower similarity scores, likely to be anomalies, receive smaller weights, reducing their impact on the model’s learning process. This strategy effectively minimizes the adverse effects of anomaly contamination, ensuring the model remains sensitive to the features of normal behavior while suppressing the noise introduced by unlabeled anomalies.

\paragraph{Clustering and Contrastive Learning}
To enhance the accuracy of anomaly detection, we propose to construct multiple normal prototypes to capture the diversity within normal data. However, initial prototypes may not fully represent the underlying structure of the normal data. Therefore, we refine these prototypes through a two-step process.

First, recognizing that the relationship between normal samples and multiple normal prototypes aligns with an unsupervised clustering problem, 
we employ Deep Embedding Clustering \cite{DeepEmbeddingClustering16} to simultaneously learn feature representations and cluster assignments. This process involves optimizing a KL divergence loss for unlabeled samples:
\begin{equation}
     \mathcal{L_{\text{kl}}}(P \parallel Q) = \sum_{i} \sum_{j} p_{ij} \log \frac{p_{ij}}{q_{ij}}
    \label{eq:kl}
\end{equation}
where 
\[
q_{ij} = \frac{s_{ij}}{\sum_{j'} s_{ij'}}, \quad f_j = \sum_{i} q_{ij}, \quad p_{ij} = \frac{q_{ij}^2 / f_j}{\sum_{j'} q_{ij'}^2 / f_{j'}}
\]
This loss function encourages the prototypes to align closely with the central tendencies of the clusters formed by the normal data, ensuring that the learned prototypes accurately reflect the true structure of the normal data distribution.

Second, to solidify the distinction between normal data and anomalous data, we introduce a contrastive learning approach. Our goal is to ensure that normal samples exhibit high similarity to at least one of the refined normal prototypes while maintaining low similarity for anomalous samples across all prototypes. This separation is enforced through a contrastive loss function:
\begin{equation}
      \mathcal{L_\text{{con}}}=-\log \sigma(\mathbb{E}_{x_i \sim \mathcal{X_U}}[w_i\max\limits_{j} s_{ij}] - \mathbb{E}_{x_i \sim \mathcal{X_A}}[\max\limits_{j} s_{ij}])
      \label{eq:con}
\end{equation}
where \(w_i\) ensures that unlabeled anomalies do not achieve high similarity to any normal prototypes.
Through this contrastive learning process, normal samples are encouraged to cluster around one of the central prototypes in the latent space, while anomalous samples are pushed to the periphery, far from any normal prototype. This method ensures a clear and robust separation between normal and anomalous data, significantly enhancing the effectiveness of the anomaly detection framework.

Finally, we combine the Clustering Loss \eqref{eq:kl} with Contrastive Loss \eqref{eq:con} to get the overall loss for this module:
\begin{equation}
    \mathcal{L_{\text{np}}} = \mathcal{L_{\text{con}}} + \mathcal{L_{\text{kl}}}
\label{eq:np}
\end{equation}

\subsection{Unified Anomaly Scoring Module}
To detect anomalies with consideration of both the reconstruction error \( e_i = \mathit{l}(x_i, \hat{x}_i) \) and the multi-normal prototype information, we design a unified anomaly scoring module. Inspired by the residual network \cite{residual17}, we also take the latent representation $z_i$ of the given sample $x_i$ as the input. In summary, the concatenated vector forms the input to the unified anomaly score evaluator \( \varphi \) to get an anomaly score:
\begin{equation}
    score_i = \varphi(\text{cat}(e_i, z_i, s_i); \Theta_{\varphi})
\end{equation}
where \( s_i = \max\limits_{j} s_{ij} \) is the maximum similarity to any normal prototype, \( \Theta_{\varphi} \) denotes the trainable parameters of the unified anomaly score evaluator, which is composed of a multi-layer perceptron with a sigmoid activation function at the final layer.

We aim for the anomaly scores to be close to 1 for anomalous samples and close to 0 for normal samples. Thus, we design the following loss function:
\begin{equation}
     \mathcal{L_{\text{score}}} = \mathbb{E}_{x_i \sim \mathcal{X_U}}[w_i \cdot \mathit{l}(score_i, 0)] + \mathbb{E}_{x_i \sim \mathcal{X_A}}[\mathit{l}(score_i, 1)]
\label{eq:score}
\end{equation}
where the weights \( w_i \) are applied to mitigate the influence of anomalies in the unlabeled data, focusing the model’s learning on the more reliable, normal-like samples.

\begin{table}[tbp]
  \caption{Overview of Datasets. \(D\) and \(N\) denote data dimension and size, respectively. $\delta$ indicates the anomaly ratio.}
  \centering
  \footnotesize
  \begin{tabular}{@{}lllll@{}}
  \toprule
  \textbf{Dataset Name}  &\textbf{Abbr.} & \(D\)  & \(N\) & \(\delta (\%)\)  \\
  \midrule
  UNSW-NB15 Dos& DoS& 196 & 109,353 & 15.0  \\
  UNSW-NB15 Rec& Rec& 196 & 106,987 & 13.1  \\
   UNSW-NB15 Backdoor& bac& 196 & 95329&2.4\\
   UNSW-NB15 Analysis& Ana& 196 & 95677&2.8\\
  IoT-23 C\&C & C\&C & 24& 31,619&17.8\\
  IoT-23 Attack & Attack & 24& 29,815&12.8\\
  IoT-23 DDoS& DDoS& 24& 26037&0.1\\
  IoT-23 Okiru& Okiru& 24& 26164&0.6\\
  Fraud Detection  & Fraud& 29 & 284,807 &0.2  \\
   Vehicle Claims & VC& 306& 268,255&21.2 \\
   Vehicle Insurance& VI& 148& 15,420&6.0 \\
   MAGIC Gamma& Gamma& 10 & 19,020 &35.2  \\
   Census Income& Census& 500 & 299,285 &6.2  \\
   Pendigits& Pendigits& 16& 6,870&2.3 \\
   Bank Marketing& Bank& 62 & 41,188 &11.3  \\
  \bottomrule
  \end{tabular}
  \label{tab:dataset}
  \end{table}

\subsection{Training and Inference}
\paragraph{Pre-training and Initialization.}
Following \cite{IDEC17}, we pre-train an AutoEncoder on unlabeled data, initially using denoising autoencoders for layer-wise training. After this, we fine-tune the entire AutoEncoder by minimizing reconstruction loss to refine the latent representations. Once fine-tuned, the encoder is used to extract latent representations of the unlabeled data. These representations are then clustered using k-means to determine $k$ cluster centers, which are used to initialize the normal prototypes. 
\paragraph{Training}
Our model is trained in an end-to-end manner using mini-batches consisting of \(b_u\) unlabeled samples and \(b_a\) labeled anomalies. For each mini-batch, we compute three losses: \(L_{recon}\), \(L_{np}\), and \(L_{score}\), as defined by Equation \eqref{eq:recon}, \eqref{eq:np}, and \eqref{eq:score}, respectively. To effectively balance these three losses dynamically, we utilize a technique known as Dynamic Averaging, inspired by \cite{DynamicAverage19}. The total loss is calculated as follows:
\begin{equation}
 \mathcal{L_{\text{total}}} = \lambda_1  \mathcal{L_\text{{recon}}} + \lambda_2  \mathcal{L_{\text{np}}} + \lambda_3 \mathcal{L_{\text{score}}}
\end{equation}
where \(\lambda_1\), \(\lambda_2\), and \(\lambda_3\) are updated per epoch based on the rate of descent of the corresponding losses.

 All parameters \((\Theta_{\{\psi,\psi^\prime\}}, \Theta_{\{u_j\}_{j=1}^k}, \Theta_{\varphi})\) are jointly optimized with respect to the total loss ensuring a comprehensive and integrated training process for the model.

\paragraph{Inference}
During the inference phase, each data instance $x_i$ is passed through the Reconstruction Learning Module to obtain its latent representation $z_i$ and reconstruction error $e_i$. The Multi-Normal Prototypes Learning Module then calculates the similarity between the latent representation and the normal prototypes, selecting the maximum similarity score  $s_i$. The three features are fed into the anomaly score evaluator \(\varphi\) to compute an anomaly score for each instance.

\section{Experiments}

\subsection{Experimental Setup}
\subsubsection{Datasets}
This section outlines the datasets used in our experiments, covering a broad range of scenarios across domains like network security, Internet of Thing (IoT) security, financial fraud detection, and others. All 15 datasets are summarized in Table \ref{tab:dataset}, providing robust benchmarks for evaluating anomaly detection methods with their diverse scenarios and varying anomaly ratios. 
The datasets include:
\begin{itemize}
    \item \textbf{UNSW-NB15}: Network traffic records with various attack types such as DoS, Reconnaissance, Backdoor, and Analysis \cite{UNSW-NB15}.
    \item \textbf{Aposemat IoT-23}: Network traffic data from IoT devices, capturing both Command and Control (C\&C) activities, direct attacks, DDoS, and Okiru \cite{garcia2020iot23}.
    \item \textbf{Credit Card Fraud Detection}: Identifies fraudulent transactions \cite{PreNet23}.
    \item \textbf{Vehicle Claims and Vehicle Insurance}: Detects anomalies in vehicle-related insurance claims \cite{chawda2022unsupervised}.
    \item \textbf{MAGIC Gamma Telescope}: Classifies high-energy gamma particles in physics and chemistry \cite{magic_gamma_telescope_159}. \hfill
    \item \textbf{Census Income}: Predicts high-income individuals \cite{census_income_20}.
    \item \textbf{Pendigits}: A pen-based handwritten digit recognition dataset \cite{pen-based_recognition_of_handwritten_digits_81}.
    \item \textbf{Bank Marketing}: Targets customer subscription predictions \cite{Moro2014ADA}.
\end{itemize}

\subsubsection{Baseline Methods}
To construct a comprehensive comparison, we adopt several different types of baseline methods. These algorithms are grouped into two categories based on their supervision level: 1) \textbf{Weakly supervised algorithms}, such as RoSAS \cite{RoSAS23}, PReNet \cite{PreNet23}, DevNet \cite{DevNet19}, DeepSAD \cite{DeepSAD20}, and FeaWAD \cite{FeaWAD21}, leverage limited labeled data alongside a larger set of unlabeled data, offering a direct comparison to our approach; 2) \textbf{Unsupervised algorithms} like DeepIForest \cite{DeepIForest23}, DeepSVDD \cite{DeepSVDD18}, and iForest \cite{IForest12}, operate without any labeled data, offering a baseline to assess the benefits of incorporating even minimal labeled data in anomaly detection. 
By comparing our method against these robust and diverse algorithms, we aim to demonstrate its effectiveness and flexibility in detecting anomalies with minimal labeled data. 

\begin{table*}[th]
  \caption{Performance comparison of different methods. The upper section presents AUC-PR (primary metric) and the lower section shows AUC-ROC. The best results are highlighted in bold and the second-best results are underlined.}
  \centering
  \scriptsize
  \resizebox{\textwidth}{!}{
    \begin{tabular}{c|ccccccccc}
  \toprule
    \textbf{Data} &\textbf{Ours}& \textbf{RoSAS}       & \textbf{PReNet}      & \textbf{DevNet}      & \textbf{DeepSAD}     & \textbf{FeaWAD}      & \textbf{DeepIForest}    &\textbf{DeepSVDD}& \textbf{IForest}   \\
  \midrule
     DoS&    \textbf{ 0.946}±0.004
  & 0.742±0.024& 0.695±0.006& \underline{0.906}±0.022& 0.489±0.184& 0.571±0.310& 0.414±0.005 &0.263±0.041& 0.148±0.007\\
    Rec&     \textbf{0.851}±0.008
  & 0.770±0.011& \underline{0.787}±0.006& 0.763±0.023& 0.695±0.051& 0.210±0.095& 0.163±0.008 &0.196±0.060& 0.104±0.004\\
    Bac&  \underline{0.870}±0.005
  & 0.543±0.023& 0.461±0.006& \textbf{0.878}±0.011& 0.509±0.036& 0.221±0.174& 0.247±0.025 &0.143±0.075&0.046±0.003\\
   Ana&  \textbf{0.910}±0.008
  & 0.556±0.045& 0.451±0.016& \underline{0.854}±0.063& 0.249±0.076& 0.268±0.235& 0.223±0.018 &0.141±0.067&0.064±0.007\\
    C\&C &     \textbf{0.519}±0.011
  & 0.367±0.026& 0.378±0.003& \underline{0.419}±0.056& 0.146±0.001& 0.247±0.061& 0.145±0.002 &0.181±0.024&0.180±0.004\\
    Attack &     \textbf{0.964}±0.012
  & \underline{0.961}±0.014& 0.906±0.008& 0.954±0.010& 0.428±0.441& 0.635±0.381& 0.162±0.004 &0.276±0.256&0.261±0.013\\
    DDoS&  \textbf{0.252}±0.202
  & 0.071±0.088& 0.003±0.002& 0.047±0.049& 0.059±0.024& 0.043±0.056& 0.117±0.014 &0.081±0.034&\underline{0.146}±0.040\\
    Okiru&  \textbf{0.518}±0.003& \underline{0.191}±0.079& 0.008±0.001& 0.144±0.090& 0.105±0.009& 0.126±0.109& 0.010±0.001 &0.012±0.016&0.019±0.001\\
    Fraud&     \textbf{0.679}±0.021
  & 0.433±0.029& 0.351±0.099& \underline{0.512}±0.152& 0.098±0.054& 0.130±0.130& 0.382±0.020 &0.021±0.022&0.139±0.034\\
    VC&     \textbf{0.697}±0.005
  & 0.497±0.006& 0.322±0.006& \underline{0.654}±0.014& 0.189±0.010& 0.256±0.037& 0.261±0.004 &0.234±0.038&0.255±0.019\\
    VI&     \textbf{0.106}±0.012
  & 0.079±0.003& 0.070±0.009& \underline{0.093}±0.005& 0.058±0.001& 0.082±0.006& 0.062±0.005 &0.063±0.004&0.064±0.002\\
    Gamma&     \underline{0.780}±0.014
  & 0.726±0.018& \textbf{0.782}±0.005& 0.749±0.013& 0.577±0.163& 0.569±0.040& 0.698±0.007 &0.437±0.021&0.644±0.008\\
    Census&     \textbf{0.392}±0.023
  & 0.309±0.018& 0.150±0.007& \underline{0.391}±0.011& 0.105±0.026& 0.183±0.086& 0.074±0.002 &0.054±0.015&0.077±0.003\\
    Pendigits& \textbf{0.856}±0.042& \underline{0.852}±0.020& 0.065±0.039& 0.766±0.060& 0.686±0.073& 0.396±0.308& 0.234±0.037 &0.073±0.088&0.274±0.025\\
    Bank&  \textbf{0.286}±0.010& 0.245±0.016& 0.146±0.012& 0.246±0.030& 0.122±0.010& 0.127±0.027& 0.233±0.006 & 0.198±0.048& \underline{0.283}±0.011\\
  \cline{1-10} 
    \textbf{Average}&  \textbf{0.642} ± 0.025& 0.489 ± 0.028& 0.372 ± 0.015& \underline{0.558} ± 0.041& 0.301 ± 0.077& 0.271 ± 0.137& 0.228 ± 0.011& 0.158 ± 0.054& 0.180 ± 0.012\\
    \textbf{P-value}& -& $<$0.0001& 0.0001& 0.0002& $<$0.0001& $<$0.0001& $<$0.0001 & $<$0.0001& $<$0.0001\\
   \bottomrule
  DoS&      \textbf{0.978}±0.003
  & 0.835±0.049& 0.795±0.006& \underline{0.963}±0.005& 0.603±0.189& 0.755±0.210& 0.787±0.015 &0.506±0.065& 0.525±0.029\\
    Rec&      \textbf{0.972}±0.003& 0.922±0.010& 0.922±0.001& \underline{0.951}±0.001& 0.820±0.057& 0.611±0.207& 0.457±0.025 &0.510±0.098& 0.410±0.021\\
    Bac&      \textbf{0.976}±0.003
  & 0.882±0.029& 0.751±0.016& \underline{0.972}±0.006& 0.660±0.033& 0.717±0.246& 0.904±0.002 &0.641±0.152& 0.727±0.009\\
    Ana&      \textbf{0.995}±0.001& 0.936±0.038& 0.782±0.022& \underline{0.987}±0.005& 0.440±0.102& 0.798±0.099& 0.916±0.004 &0.564±0.122&0.781±0.022\\
    C\&C &      \textbf{0.733}±0.003
  & 0.723±0.001& \underline{0.730}±0.003& 0.729±0.008& 0.427±0.002& 0.568±0.057& 0.426±0.007 &0.492±0.039&0.495±0.004\\
    Attack &      \textbf{0.999}±0.000& \underline{0.998}±0.002& 0.992±0.000& \underline{0.998}±0.000& 0.410±0.481& 0.815±0.281& 0.643±0.010 &0.397±0.461&0.820±0.013\\
    DDoS&  \textbf{0.822}±0.163
  & 0.427±0.066& 0.521±0.067& 0.466±0.079& 0.673±0.042& 0.494±0.124& \underline{0.783}±0.087 &0.504±0.067&0.780±0.002\\
    Okiru&  \underline{0.960}±0.001& 0.951±0.014& 0.579±0.028& \textbf{0.961}±0.005& 0.946±0.012& 0.896±0.109& 0.692±0.026 &0.676±0.323&0.845±0.009\\
    Fraud&      \underline{0.964}±0.006
  & 0.896±0.015& 0.867±0.032& \textbf{0.967}±0.015& 0.744±0.105& 0.655±0.280& 0.962±0.002 &0.533±0.094&0.957±0.003\\
    VC&      \textbf{0.774}±0.005
  & 0.702±0.007& 0.559±0.004& \underline{0.754}±0.005& 0.373±0.016& 0.544±0.029& 0.559±0.005 &0.451±0.037&0.561±0.026\\
    VI&      \textbf{0.693}±0.031& 0.578±0.013& 0.528±0.040& \underline{0.639}±0.016& 0.476±0.008& 0.606±0.030& 0.486±0.030 &0.502±0.015&0.501±0.012\\
    Gamma&      \underline{0.845}±0.008
  & 0.807±0.014& \textbf{0.846}±0.003& 0.837±0.004& 0.663±0.156& 0.700±0.057& 0.773±0.008 &0.536±0.017&0.727±0.007\\
    Census&      \textbf{0.896}±0.003& 0.781±0.026& 0.574±0.007& \underline{0.895}±0.004& 0.429±0.059& 0.703±0.172& 0.614±0.010 &0.463±0.078&0.627±0.015\\
    Pendigits&      \underline{0.986}±0.007& \textbf{0.989}±0.002& 0.710±0.121& 0.971±0.014& 0.823±0.113& 0.761±0.214& 0.949±0.008 &0.770±0.202&0.953±0.005\\
    Bank&  \textbf{0.701}±0.008& 0.676±0.015& 0.552±0.025& 0.677±0.019& 0.470±0.022& 0.516±0.087& 0.666±0.003 &0.541±0.036&\underline{0.691}±0.012\\
  \cline{1-10} 
    \textbf{Average}&  \textbf{0.886} ± 0.016& 0.807 ± 0.020& 0.714 ± 0.025& \underline{0.851} ± 0.012& 0.597 ± 0.093& 0.676 ± 0.147& 0.708 ± 0.016& 0.539 ± 0.120& 0.693 ± 0.013\\
    \textbf{P-value}& -& 0.0002& 0.0001& 0.0009& $<$0.0001& $<$0.0001& $<$0.0001 & $<$0.0001& $<$0.0001\\
  \bottomrule
    \end{tabular}
  }
  \label{tab:Performance Comparison}
  \end{table*}

\subsubsection{Evaluation Metrics}
The same as previous studies \cite{RoSAS23,PreNet23,DevNet19}, we primarily use the Area Under the Precision-Recall Curve (AUC-PR) as our main evaluation metric, supplemented by the Area Under the Receiver Operating Characteristic Curve (AUC-ROC). The AUC-PR is particularly effective for imbalanced datasets, emphasizing the model's ability to correctly identify anomalies (true positives) while minimizing false positives and false negatives, making it ideal for anomaly detection where anomalies are rare. The AUC-ROC provides an aggregate measure of the model's performance across all classification thresholds, illustrating its ability to distinguish between positive and negative classes. These two metrics both range from 0 to 1 and higher values indicate better performance. To ensure robust and reliable results, we report the average performance over 5 independent runs and employ the paired Wilcoxon signed-rank test\cite{doi:https://doi.org/10.1002/9780471462422.eoct979} to statistically validate the significance of our findings.

\subsubsection{Implementation Details}
We use the Adam optimizer with a learning rate of \(5 \times 10^{-3}\) and weight decay of \(5 \times 10^{-4}\). The latent representation dimension \(H\) is set to 8, and the hyperparameters are configured as follows: batch sizes \(b_u = 128\), \(b_a = 32\); margin \(m_1 = 0.02\); Student’s t-distribution degree of freedom \(\alpha = 1\); and similarity scaling parameter \(\beta = 1\). The number of normal prototypes k is determined via clustering analysis on the unlabeled samples. The implementation of the compared methods is primarily based on the DeepOD framework\cite{DeepIForest23} and \texttt{scikit-learn} (\texttt{sklearn}).

\subsection{Performance Comparison}

The same as in \cite{weakly-survey23}, we evaluate the performance of our proposed anomaly detection method across multiple datasets with a fixed labeled anomaly ratio of 1.0\% for each dataset. The labeled anomaly ratio is defined as the proportion of anomalies in the training dataset that are labeled. 
This means that in practical anomaly detection scenarios, we only need to label anomalies included in 1.0\% of the training data, ensuring that the weakly supervised anomaly detection setup does not affect the proportion of anomalies in the training set regardless of the inherent anomaly ratio of the dataset.

Table \ref{tab:Performance Comparison} shows the performance of different methods on the fifteen benchmark datasets. Our method consistently outperformed other approaches, achieving the highest average scores in both AUC-PR and AUC-ROC metrics, with statistically significant improvements with over 99.9\% confidence. Specifically, in terms of AUC-PR, our method achieved the best results on 13 of the 15 datasets, and was the second-best on the remaining two, closely approaching the top results. The average AUC-PR of our method was 0.642, which is significantly higher than other methods. Similarly, for the AUC-ROC metric, our method achieved an average of 0.886, surpassing all other methods. The statistical validation further confirmed that the improvements brought by our method are significant, providing robust and reliable anomaly detection performance in weakly supervised settings. 

Besides, the weakly supervised algorithms, such as our method, RoSAS, and DevNet, perform significantly better than all unsupervised algorithms on almost all datasets. This confirms that with the appropriate use of a few labeled anomaly samples, we can get better anomaly detection performance. 

\begin{figure*}[th]
  \centering
  \includegraphics[width=0.8\textwidth]{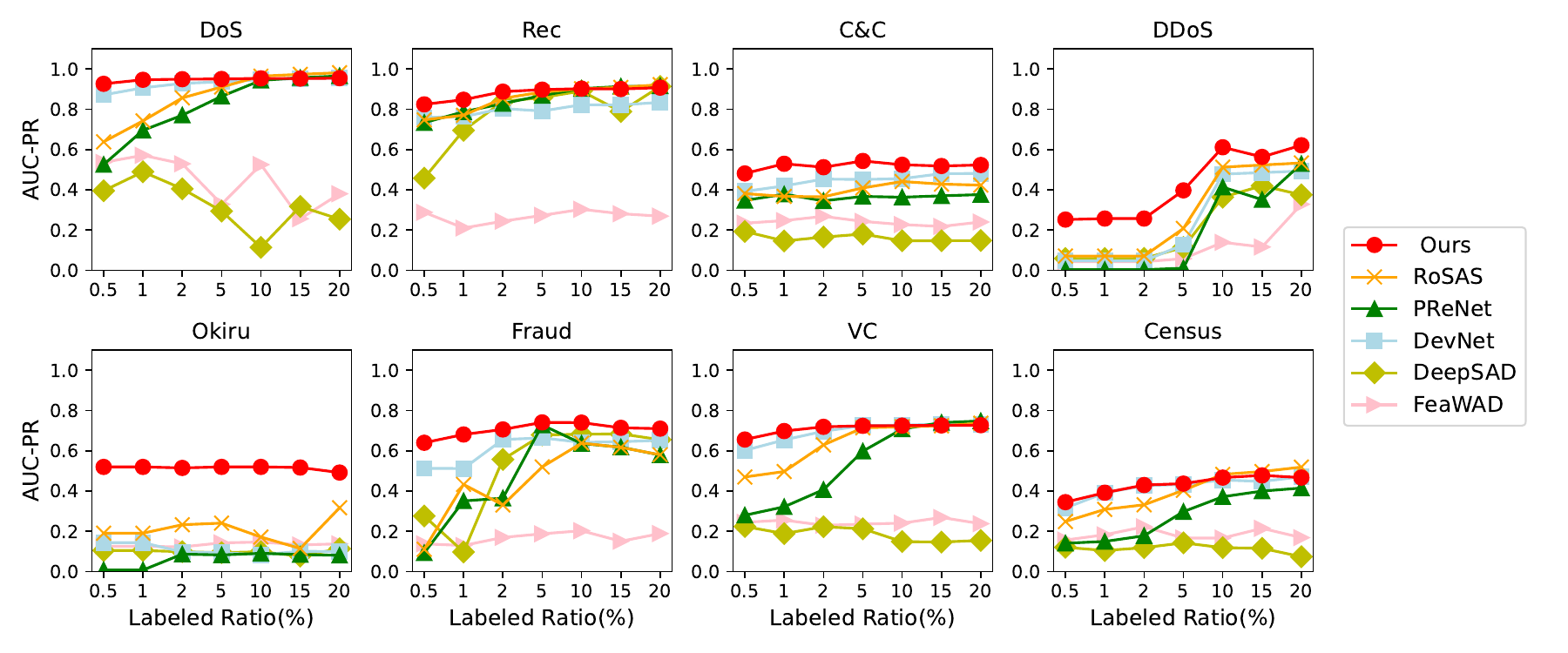}
  \caption{Effects of varying the ratios of labeled anomalies.}
  \label{fig:Labeled Anomaly ratios}
\end{figure*}

\begin{table*}[h]
\caption{ Detection of unseen anomalies, with performance measured by AUC-PR. The "Seen" indicates the anomaly type included in the training and validation sets, while the "Unseen" indicates the anomaly type only included in the test set.}
\centering
\scriptsize
\resizebox{\textwidth}{!}{
  \begin{tabular}{ll|ccccccccc}
\toprule
 \textbf{Seen}&\textbf{Unseen}&\textbf{Ours}& \textbf{RoSAS}& \textbf{PReNet}& \textbf{DevNet}& \textbf{DeepSAD}& \textbf{FeaWAD}& \textbf{DeepIForest}  &\textbf{DeepSVDD}& \textbf{IForest}\\
\midrule
  Rec&DoS&        \textbf{0.824}±0.041& 0.515±0.042& 0.354±0.045& 0.512±0.067& 0.311±0.092& 0.279±0.146& 0.654±0.013 &0.311±0.083& 0.212±0.013\\
  Ana&DoS&        \textbf{0.904}±0.011& 0.710±0.053& 0.546±0.019& 0.882±0.015& 0.415±0.080& 0.651±0.180& 0.659±0.014 &0.370±0.121& 0.248±0.036\\
  Bac&DoS&        \textbf{0.868}±0.009& 0.654±0.060& 0.538±0.012& 0.867±0.009& 0.461±0.069& 0.658±0.245& 0.668±0.024 &0.312±0.077& 0.256±0.030\\
 DoS& Rec& \textbf{0.681}±0.033& 0.653±0.022& 0.389±0.040& 0.566±0.066& 0.324±0.104& 0.274±0.145& 0.161±0.004 &0.199±0.042&0.097±0.007\\
 Ana& Rec& \textbf{0.534}±0.105& 0.250±0.026& 0.176±0.004& 0.469±0.047& 0.257±0.063& 0.281±0.038& 0.210±0.011 &0.275±0.080&0.133±0.005\\
 Bac& Rec& \textbf{0.699}±0.057& 0.566±0.048& 0.478±0.037& 0.535±0.031& 0.375±0.011& 0.297±0.120& 0.213±0.007 &0.197±0.054&0.124±0.007\\
 DoS& Ana& \textbf{0.844}±0.015& 0.537±0.010& 0.493±0.004& 0.843±0.016& 0.266±0.110& 0.477±0.243& 0.116±0.009 &0.097±0.043&0.034±0.004\\
 Rec& Ana& \textbf{0.586}±0.033& 0.180±0.019& 0.082±0.013& 0.198±0.051& 0.138±0.067& 0.077±0.047& 0.253±0.020 &0.137±0.068&0.056±0.004\\
 Bac& Ana& 0.785±0.020& 0.417±0.034& 0.365±0.009& \textbf{0.828}±0.021& 0.324±0.062& 0.244±0.243& 0.263±0.006 &0.231±0.114&0.069±0.010\\
  DoS&Bac&        \textbf{0.865}±0.018& 0.615±0.016& 0.578±0.007& 0.851±0.013& 0.313±0.127& 0.525±0.267& 0.105±0.006 &0.085±0.048&0.021±0.001\\
  Rec&Bac&       \textbf{0.735}±0.023& 0.320±0.055& 0.169±0.049& 0.166±0.021& 0.153±0.059& 0.095±0.036& 0.250±0.030 &0.226±0.126&0.045±0.009\\
  Ana&Bac&        \textbf{0.857}±0.004& 0.415±0.045& 0.334±0.021& 0.815±0.068& 0.289±0.079& 0.238±0.239& 0.243±0.016 &0.153±0.038&0.051±0.007\\
\hline
 \multicolumn{2}{c}{\textbf{Average}}&\textbf{0.765} ± 0.031& 0.486 ± 0.036& 0.375 ± 0.022& 0.628 ± 0.035& 0.302 ± 0.077& 0.341 ± 0.162& 0.316 ± 0.013 &0.216 ± 0.075& 0.112 ± 0.011\\
 \multicolumn{2}{c}{\textbf{P-value}}& -& 0.0005&  0.0005& 0.0068& 0.0005&  0.0005& 0.0005 &0.0005 & 0.0005\\
 \bottomrule
  \end{tabular}
  }

\label{tab:unseen detection}
\end{table*}

\subsection{Effects of More or Less Labeled Anomalies}
This experiment aims to examine the effects of Labeled Anomaly Ratios on the model's effectiveness in detecting anomalies, helping us understand the trade-off between supervision level and detection performance. 
The Labeled Anomaly Ratios refer to the proportion of labeled anomaly samples within the dataset. 
A higher ratio implies more known information about anomalies, bringing the learning process closer to a supervised setting and potentially improving the model's anomaly detection capabilities, albeit at a higher labeling cost. Conversely, a lower ratio indicates less labeled information and lower labeling costs but challenges the model's detection abilities due to reduced guidance. Because AUC-ROC and AUC-PR show almost consistent results, we only present AUC-PR results here. Additionally, due to space limitations, we only showcase a subset of datasets. In Fig. \ref{fig:Labeled Anomaly ratios}, the $x$ axis represents Labeled Anomaly Ratios ranging from 0.5\% to 20\%, and the $y$ axis represents the AUC-PR values for each dataset across all methods (except for unsupervised methods, as they are not affected by the labeled anomaly ratios).

From Fig. \ref{fig:Labeled Anomaly ratios}, it is evident that our method consistently achieves the best performance across nearly all Labeled Anomaly Ratios. Specifically, as the Labeled Anomaly Ratios decrease, our method's performance degrades more slowly and maintains relatively high detection performance, while other methods exhibit a significant decline in most datasets. This indicates that our method does not overly rely on the small fraction of labeled anomalies and effectively utilizes the modeling of multiple normal prototypes in a large amount of unlabeled data.

On the other hand, from the perspective of anomaly contamination, as the proportion of labeled anomalies decreases, the proportion of unlabeled anomalies contaminating the unlabeled data increases. Fig. \ref{fig:Labeled Anomaly ratios} shows that, unlike other methods, our method's detection performance declines more slowly as the anomaly contamination ratio increases, maintaining a relatively good detection performance. This demonstrates the robustness of our model against anomaly contamination. This robustness is attributed to the model’s design, which estimates the likelihood of each unlabeled sample being normal and uses this estimation as a weight to reduce the impact of contaminating anomalies.

\begin{table*}[th]
  \caption{Ablation study results on 15 datasets. Each column represents the removal of a specific component from the full model. The average AUC-PR and P-values are reported to demonstrate the contribution of each component.}
  \centering
  \scriptsize
    \begin{tabular}{c|cccccc}
  \toprule
    \textbf{Data} &\textbf{Full Model}& \textbf{w/o Pretraining}       & \textbf{w/o \(\mathcal{L_{\text{recon}}}\)}      & \textbf{w/o Decoder}      &\textbf{w/o \(w_i\)}      & \textbf{w/o MNP}     \\
  \midrule
     DoS&    \textbf{ 0.946}±0.004
  & 0.938±0.014& 0.943±0.005& 0.942±0.004 &0.945±0.003& 0.918±0.019\\
    Rec&     0.851±0.008
  & 0.840±0.015& 0.852±0.012&  \textbf{0.855}±0.014 &0.836±0.022& 0.816±0.014\\
    Bac&  \textbf{0.870}±0.005& 0.842±0.045& 0.847±0.009&  0.833±0.039 &0.865±0.013& 0.802±0.066\\
   Ana&  \textbf{0.910}±0.008
  & 0.885±0.048& 0.864±0.065&  0.855±0.074 &0.830±0.064& 0.805±0.015\\
    C\&C &     \textbf{0.519}±0.011
  & 0.488±0.044& 0.465±0.061&  0.506±0.019 &0.498±0.046& 0.476±0.036\\
    Attack &     \textbf{0.964}±0.012
  & 0.894±0.089& 0.952±0.004&  0.959±0.015 &0.963±0.011& 0.875±0.030\\
    DDoS&  \textbf{0.252}±0.202
  & 0.161±0.054& 0.206±0.170&  0.243±0.291 &0.242±0.193& 0.128±0.023\\
    Okiru&  0.518±0.003& 0.425±0.190& 0.457±0.122&  \textbf{0.520}±0.000 &0.314±0.220& \textbf{0.520}±0.000\\
    Fraud&     \textbf{0.679}±0.021
  & 0.663±0.063& 0.649±0.027&  0.611±0.082 &0.665±0.017& 0.601±0.012\\
    VC&     0.697±0.005
  & 0.680±0.006& 0.689±0.009&  0.697±0.008 &\textbf{0.699}±0.003& 0.633±0.007\\
    VI&     \textbf{0.106}±0.012
  & 0.087±0.007& 0.099±0.004&  0.098±0.010 &0.104±0.010& 0.088±0.008\\
    Gamma&     \textbf{0.780}±0.014& 0.770±0.007& 0.767±0.016&  0.761±0.029 &0.768±0.013& 0.680±0.016\\
    Census&     \textbf{0.392}±0.023
  & 0.380±0.012& 0.389±0.016&  0.371±0.009 &0.374±0.016& 0.341±0.017\\
    Pendigits& \textbf{0.856}±0.042& 0.844±0.055& 0.829±0.070&  0.845±0.061 &0.821±0.026& 0.803±0.067\\
    Bank&  \textbf{0.286}±0.010& 0.252±0.038& 0.227±0.020&  0.232±0.014 &0.277±0.014& 0.231±0.020\\
  \cline{1-7} 
    \textbf{Average}&  \textbf{0.642} ± 0.025& 0.610 ± 0.046& 0.616 ± 0.041&  0.622 ± 0.045&0.614 ± 0.045& 0.581 ± 0.023\\
    \textbf{P-value}& -& $<$0.0001& 0.0001& 0.0017&0.0034& $<$0.0001\\
   \bottomrule
    \end{tabular}
  \label{tab:ablation study}
  \end{table*}

\subsection{Detection Performance on Unseen Anomalies}
A crucial aspect of an effective anomaly detection system is its ability to generalize from known anomalies to detect previously unseen ones. This capability is essential for maintaining robustness in dynamic environments where new types of anomalies may emerge. To evaluate our method’s performance on unseen anomalies, we constructed several experimental datasets using the UNSW-NB15 dataset, which contains four types of anomalies: DoS, Rec, Ana, and Bac. In each experiment, two types of anomalies were selected: one used exclusively in the test set (unseen) and the other included in the training and validation sets (seen). 
Consistent with our weakly supervised anomaly detection settings, the labeled anomaly ratio in the training set was maintained at 1.0\%. This setup resulted in 12 different experimental datasets.
 
The results are presented in Table \ref{tab:unseen detection}, showing the AUC-PR scores for different methods across various combinations of seen and unseen anomaly types.  
Our method achieved the highest average AUC-PR of 0.765, significantly outperforming other methods by more than 20\%. The statistical confidence level for this improvement exceeds 99.0\%.
This experiment demonstrates that our method can effectively generalize from known anomalies to detect previously unseen ones, maintaining strong detection performance and robustness in dynamic environments. 
The success of our approach may be attributed to its ability to maintain multiple normal prototypes, which better represent the diverse nature of normal behavior. This focus on distinguishing data points based on their contrast with the prototypes, rather than relying exclusively on previously limited labeled anomalies, likely contributes to the observed improvements.

\begin{figure}[t]
  \centering
  \includegraphics[width=0.9\columnwidth]{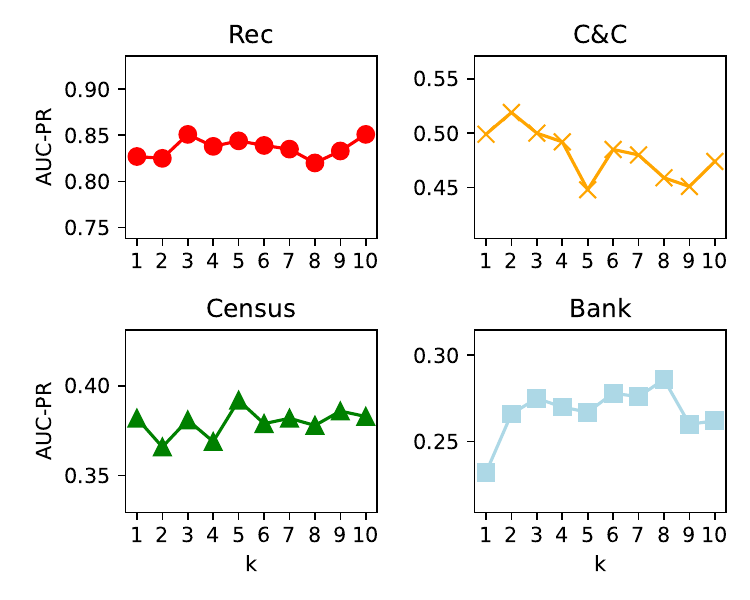}
  \caption{Effects of varying the number of normal prototypes}
  \label{fig: Normal Prototype}
\end{figure}

\subsection{Ablation Study}

We conducted an ablation study to evaluate the contribution of each component in our model. Table \ref{tab:ablation study} summarizes the results across 15 datasets. The full model achieves the highest average AUC-PR of \(0.642\), demonstrating the effectiveness of the complete framework. Below, we analyze the impact of removing each component in order.

\textbf{w/o Pretraining:} Removing the pretraining step generally decreases performance, as it negatively impacts the model initialization. Overall, the effect of pretraining is more pronounced in smaller datasets, while its impact diminishes as dataset size increases. For instance, in the relatively small Aposemat IoT-23 datasets such as DDoS (\(0.252 \pm 0.202 \to 0.161 \pm 0.054\)) and Okiru (\(0.518 \pm 0.003 \to 0.425 \pm 0.190\)), pretraining significantly boosts performance. In contrast, larger datasets such as UNSW-NB15 DoS (\(0.946 \pm 0.004 \to 0.938 \pm 0.014\)) show a relatively smaller degradation when pretraining is removed. These results indicate that pretraining provides essential initialization benefits, particularly for datasets with limited size. Although the degradation is moderate, pretraining proves beneficial across diverse scenarios, enhancing the robustness of the model.

\textbf{w/o \(\mathcal{L_{\text{recon}}}\):} The reconstruction loss \( \mathcal{L_{\text{recon}}} \) is crucial for learning representations of normal data. Removing this component results in significant performance degradation in datasets like Ana (\(0.910 \pm 0.008 \to 0.864 \pm 0.065\)) and C\&C (\(0.519 \pm 0.011 \to 0.465 \pm 0.061\)). These results demonstrate that reconstruction loss helps model normal patterns and separate them from anomalies effectively. However, its absence has a smaller effect on datasets like Rec, where the AUC-PR remains stable (\(0.851 \pm 0.008 \to 0.852 \pm 0.012\)).



\begin{figure*}[th] %
  \centering
  \subfigure[Synthetic Multi-Modal Data.]{
      \includegraphics[width=0.25\textwidth]{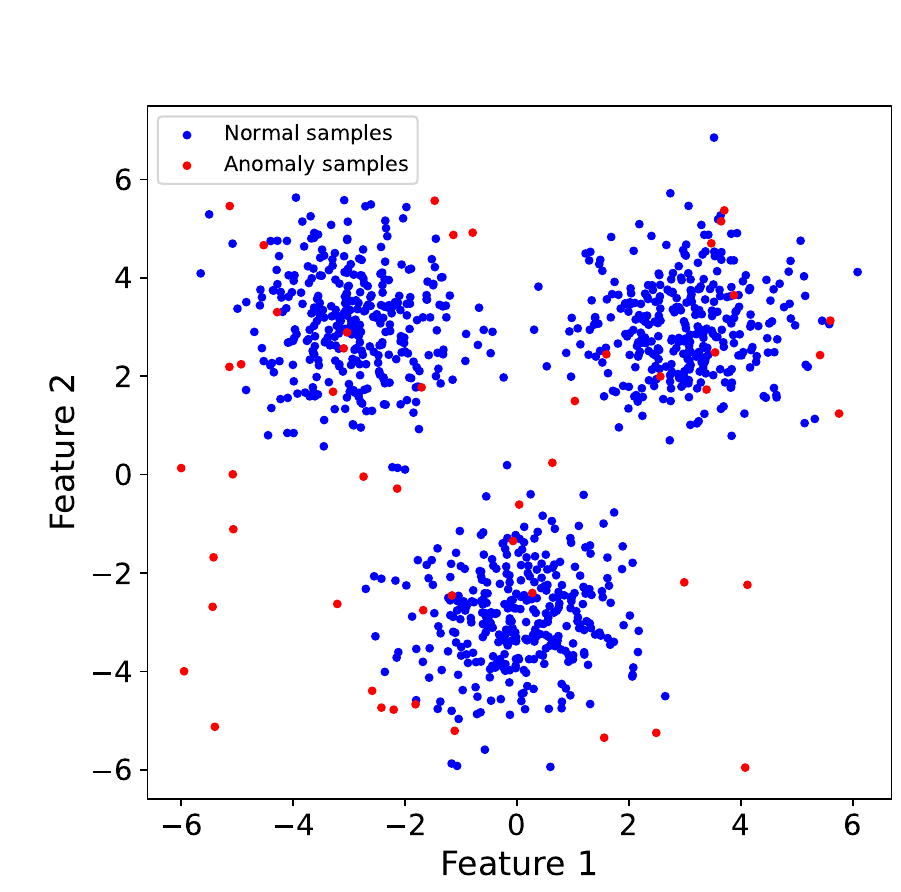} %
      \label{fig:subfig1}
  }
  \hspace{5mm}
  \subfigure[Multi-normal Prototypes Anomaly Detection.]{
      \includegraphics[width=0.33\textwidth]{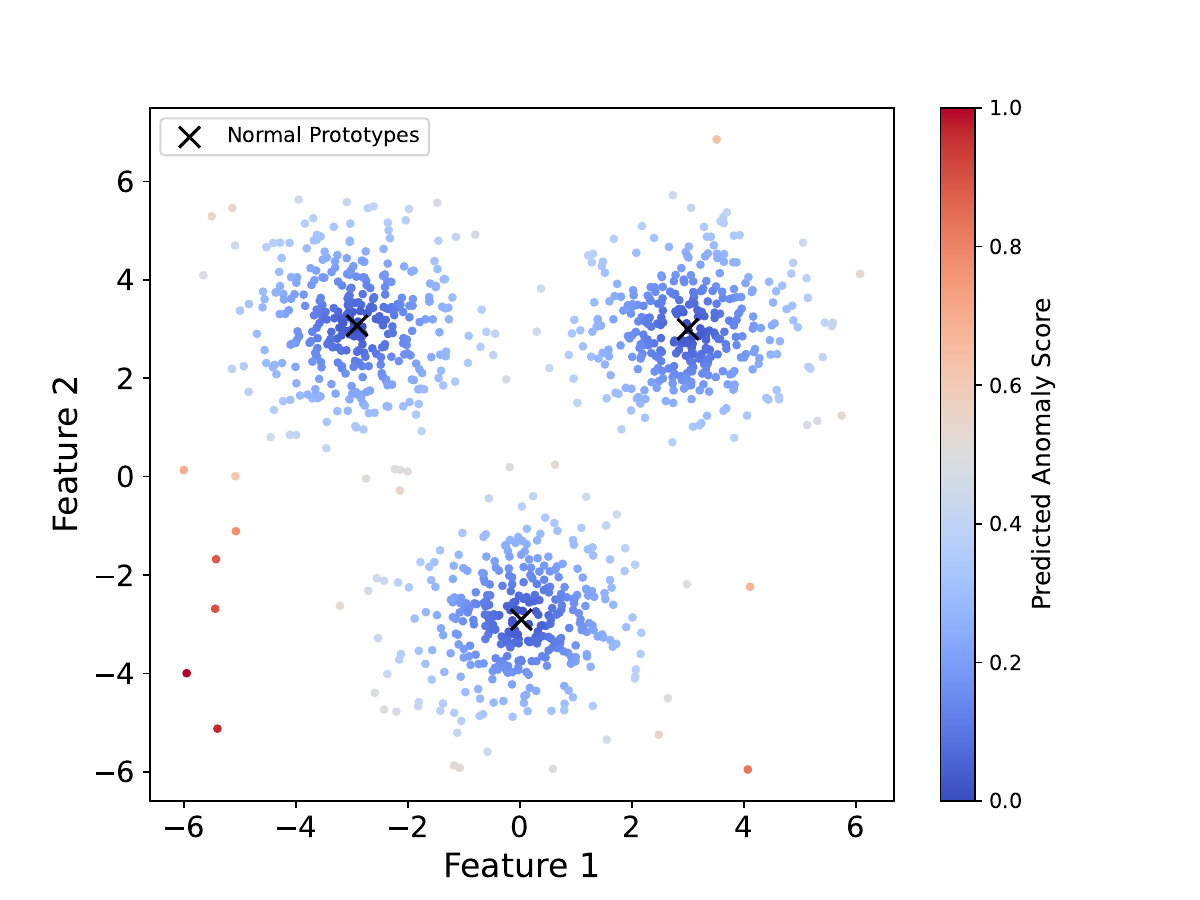} %
      \label{fig:subfig2}
  }
  \hfill
  \subfigure[Single-normal Prototypes Anomaly Detection.]{
      \includegraphics[width=0.33\textwidth]{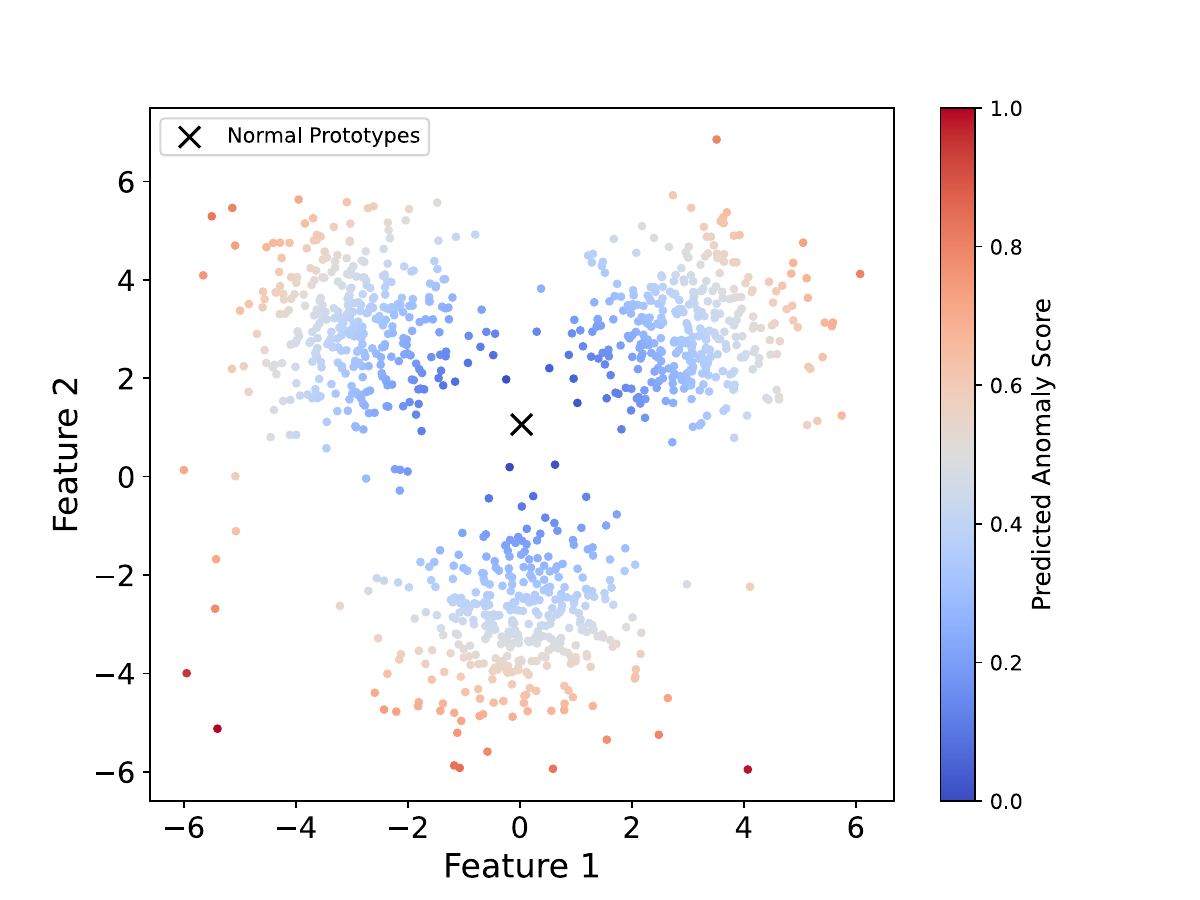} %
      \label{fig:subfig3}
  }
  \hfill

  \caption{Anomaly detection results on synthetic data. (a) The synthetic dataset comprises three Gaussian clusters representing normal samples and scattered anomalies. (b) Anomaly detection using multi-normal prototypes successfully captures the three distinct normal modes, assigning low anomaly scores to normal samples while effectively identifying anomalies. (c) Anomaly detection using a single normal prototype fails to adapt to the multi-modal nature of the normal data, resulting in misclassified normal samples and missed anomalies.}
  \label{fig:Synthetic Normal Prototype}
\end{figure*}

\textbf{w/o Decoder:} Removing the decoder (and thus the reconstruction error) shows varied effects. In the Okiru dataset, the AUC-PR remains unchanged (\(0.518 \pm 0.003\)), suggesting that the decoder plays a less significant role for simpler data structures. However, on the Fraud dataset, the AUC-PR drops substantially (\(0.679 \pm 0.021 \to 0.611 \pm 0.082\)), highlighting its importance for datasets with highly imbalanced distributions.

\textbf{w/o \(w_i\):} The sample weighting mechanism \(w_i\) is designed to dynamically adjust the contributions of samples, particularly to address noise, imbalance, and ambiguity in the training process. Its removal results in varying levels of degradation depending on the dataset characteristics. For example, in the Okiru dataset (\(0.518 \pm 0.003 \to 0.314 \pm 0.220\)), the absence of \(w_i\) leads to a substantial decline, highlighting its role in filtering noisy or ambiguous samples. Even in datasets with moderate imbalance, such as Rec (\(0.851 \pm 0.008 \to 0.836 \pm 0.022\)), removing \(w_i\) reduces performance, demonstrating its utility in enhancing the model's robustness. 

\textbf{w/o MNP:} The exclusion of multi-normal prototypes (MNP) leads to the most significant performance degradation across almost all datasets, confirming its critical role in the framework. Notably, the Attack (\(0.964 \pm 0.012 \to 0.875 \pm 0.030\)) and DDoS (\(0.252 \pm 0.202 \to 0.128 \pm 0.023\)) datasets exhibit the largest drops, further demonstrating that single-prototype representations fail to capture the multi-modal nature of normal data.

\subsection{Impact of Normal Prototypes: Empirical and Synthetic Analysis}

In this section, we examine the effect of varying the number of normal prototypes through both empirical studies on real-world datasets and controlled experiments on synthetic data.

As illustrated in Fig. \ref{fig: Normal Prototype}, which presents results on four representative real-world datasets, the number of prototypes significantly influences the anomaly detection performance. For most datasets, using multiple prototypes yields better results, supporting our hypothesis that normal data often exhibits multi-modal characteristics. In the Rec dataset, the prototype count has minimal effect, suggesting lower sensitivity to this parameter. However, in the C\&C and Census datasets, the number of prototypes plays a critical role—both too many and too few prototypes lead to substantial performance degradation. In the Bank dataset, despite fluctuations in performance with varying prototypes, the results consistently surpass those achieved with a single prototype. 

To further validate the multi-modal nature of normal data and the effectiveness of multiple prototypes, we conducted experiments on a synthetic dataset designed with explicit multi-modal characteristics. The synthetic dataset consists of three Gaussian clusters representing normal samples and a small number of uniformly distributed anomalies, as shown in Fig. \ref{fig:Synthetic Normal Prototype} (a). We compare the anomaly detection results using multi-normal prototypes and a single normal prototype. Fig. \ref{fig:Synthetic Normal Prototype} (b) demonstrates the effectiveness of multi-normal prototypes in accurately capturing the three distinct modes of the normal data distribution. The model assigns low anomaly scores to normal samples within their respective clusters while effectively identifying anomalies scattered across the feature space. In contrast, Fig. \ref{fig:Synthetic Normal Prototype} (c) shows the results with a single normal prototype. Here, the model fails to adapt to the multi-modal distribution of the normal data, resulting in higher anomaly scores even for normal samples located farther from the single prototype. Additionally, anomalies positioned in the central region, equidistant from the three normal modes, are often missed due to their proximity to the single prototype. This failure illustrates that a single prototype cannot adequately represent diverse normal patterns, leading to poorer anomaly detection performance.

\begin{figure}[t]
  \centering
  \includegraphics[width=0.9\columnwidth]{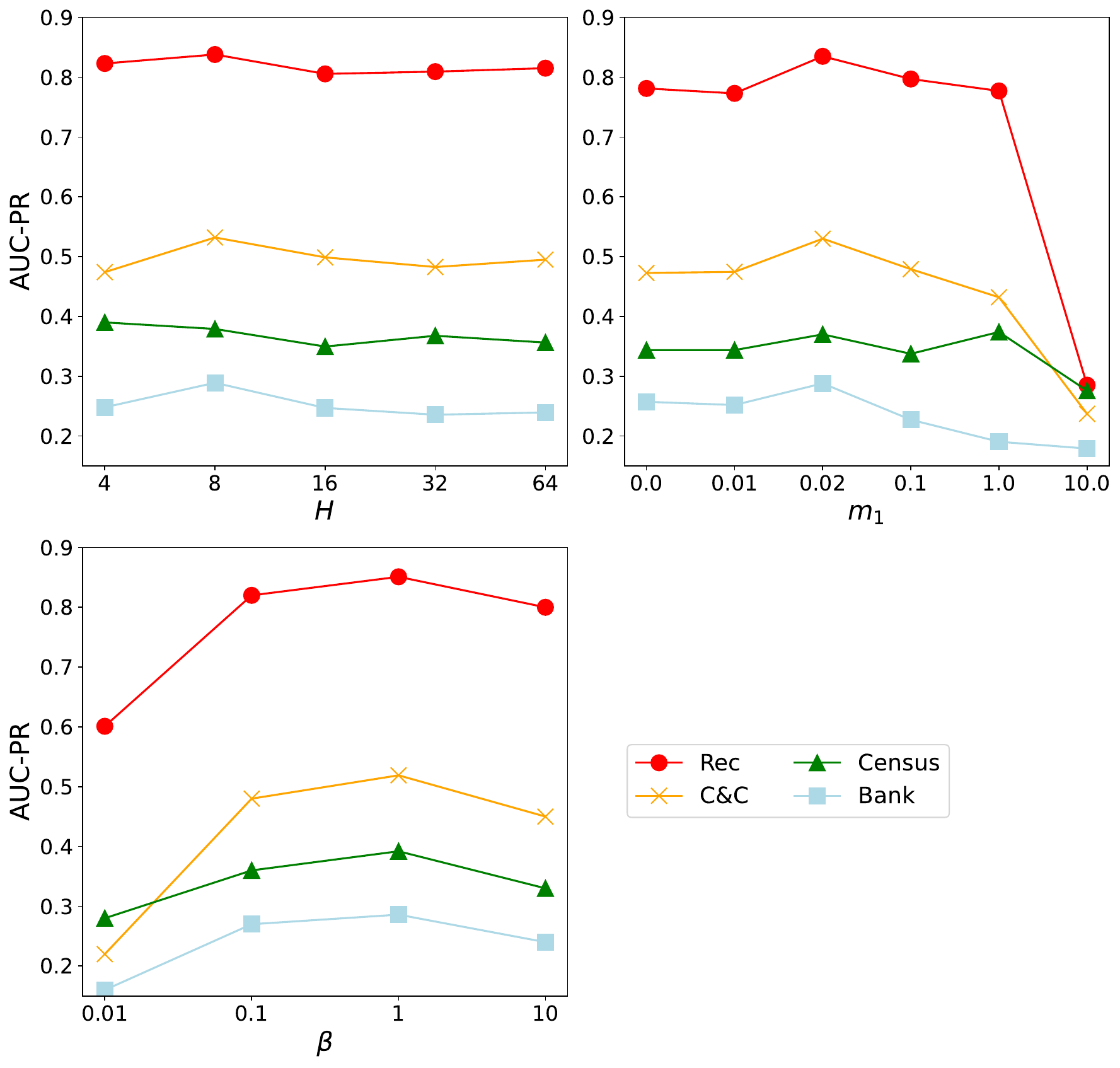}
  \caption{AUC-PR of our method w.r.t. Three key hyperparameters}
  \label{fig: sensitive}
\end{figure}

\subsection{Sensitivity Analysis}

We assess the model’s sensitivity to three key hyperparameters: the latent representation dimension \( H \), the margin \( m_1 \), and the scaling parameter \( \beta \). As shown in Fig. \ref{fig: sensitive}, we analyze the trends by selecting four representative datasets (Rec, C\&C, Census, and Bank).

For \( H \), the model shows relative stability across different values, with \( H = 8 \) providing a good trade-off between performance and computational complexity. Increasing \( H \) beyond 8 does not result in significant performance gains and may introduce unnecessary computational overhead. The consistent performance across \( H \) values highlights the model’s ability to effectively capture the underlying data structure even with relatively small latent dimensions.

For \( m_1 \), a margin value of approximately \( 0.02 \) achieves the best performance across all datasets. Smaller margins, such as \( m_1 = 0.01 \), fail to adequately separate the reconstruction losses of normal and anomalous samples, leading to reduced detection accuracy. On the other hand, larger margins, particularly \( m_1 = 1.0 \) and \( m_1 = 10.0 \), hinder the model’s capacity to learn anomaly-specific latent representations, resulting in substantial performance degradation.

For \( \beta \), which controls the scaling of dynamic weights, the optimal value is around \( \beta = 1 \). Smaller values (e.g., \( \beta = 0.1 \)) reduce the effect of dynamic weighting, limiting the model’s ability to suppress noisy samples, while larger values (e.g., \( \beta = 10 \)) overemphasize specific samples, potentially causing overfitting or instability. The improvements observed around \( \beta = 1 \) demonstrate its critical role in balancing sample contributions effectively during training.

\section{Conclusions}
In conclusion, our proposed anomaly detection method demonstrates significant improvements over existing techniques by effectively leveraging a small amount of labeled anomaly data alongside a large amount of unlabeled data. Our method's robustness to anomaly contamination and its ability to generalize to unseen anomalies make it highly suitable for real-world applications across various domains, including network security\cite{cybersecurity24}, financial fraud detection\cite{FinancialIoT24}, and medical diagnostics\cite{healthcare24}. Future work could explore extending our method to other types of data, such as images, to further validate its versatility and effectiveness. For the reconstruction learning component, we could also consider using more advanced variants like Variational AutoEncoders (VAEs)\cite{VAE13} to enhance feature extraction capabilities. 

\bibliographystyle{IEEEtran}
\bibliography{IEEEabrv,reference}
\end{document}